\documentclass{ecai}

\usepackage{cite}
\usepackage{url}
\usepackage{float}
\usepackage{amsmath,amssymb,amsfonts}
\usepackage{algorithmic}
\usepackage{textcomp}
\usepackage{xcolor}

\usepackage{times}
\usepackage{graphicx}
\usepackage{latexsym}
\usepackage{hyperref}

\DeclareMathOperator*{\argmin}{arg\,min}

\makeatletter
\def\blfootnote{\xdef\@thefnmark{}\@footnotetext}
\makeatother

\begin{document}

\title{Bias in Machine Learning - \\
What is it Good for?}

\author{Thomas Hellstr\"om \and Virginia Dignum \and Suna Bensch \institute{Ume{\aa} University,
Sweden, email: \{thomas.hellstrom,virginia.dignum,\newline suna.bensch\}@umu.se} \textsuperscript{,}\footnote{This research is partly funded by the Swedish funding agency VINNOVA, project number 2019-01930.}}

\maketitle

\begin{abstract}
  In public media as well as in scientific publications, the term \emph{bias} is used in conjunction with machine learning in many different contexts, and with many different meanings. This paper proposes a taxonomy of these different meanings, terminology, and definitions by surveying the, primarily scientific, literature on machine learning. In some cases, we suggest extensions and modifications to promote a clear terminology and completeness. The survey is followed by an analysis and discussion on how different types of biases are connected and depend on each other. We conclude that there is a complex relation between  bias occurring in the machine learning pipeline that leads to a model, and the eventual bias of the model (which is typically related  to social discrimination). The former bias may or may not influence the latter, in a sometimes bad, and sometime good way.
  
\end{abstract}

\section{INTRODUCTION}
Media, as well as scientific publications, frequently report on `Bias in Machine Learning', and how systems based on AI or machine learning are `sexist' 
\footnote{\href{https://www.bbc.com/news/technology-45809919}{BBC News, Oct. 10, 2018}. Accessed July. 29, 2020. }
or `discriminatory'
\footnote{\href{https://www.reuters.com/article/us-amazon-com-jobs-automation-insight/amazon-scraps-secret-ai-recruiting-tool-that-showed-biasagainst-women-idUSKCN1MK08G}{Reuters Technology News, Oct. 10, 2018}. Accessed July 29, 2020.}\cite{Chouldechova2016FairPW,Pedreshi08}.
\blfootnote{Copyright © 2020 for this paper by its authors. Use permitted under Creative Commons License Attribution 4.0 International (CC BY 4.0). }
In the field of machine learning, the term bias has an established historical meaning that, at least on the surface, totally differs from how the term is used in typical news reporting. Furthermore, even within machine learning, the term is used in very many different contexts and with very many different meanings. Definitions are not always given, and if they are, the relation to other usages of the word is not always clear. Furthermore, definitions sometimes overlap or contradict each other~\cite{campolo2018ai}.

The main contribution of this paper is a proposed taxonomy of the various meanings of the term bias in conjunction with machine learning. When needed, we suggest extensions and modifications to promote a clear terminology and completeness.
We argue that this is more than a matter of definitions of terms. Terminology shapes how we identify and approach problems, and furthermore how we communicate with others. This is particularly important in multidisciplinary work, such as application-oriented machine learning.

The taxonomy is based on a survey of published research in several areas, and is followed by a discussion on how different types of biases are connected and depend on each other.

Since humans are involved in both the creation of bias, and in the application of, potentially biased, systems, the presented work is related to several of the AI-HLEG recommendations for building Human-Centered AI systems.

Machine learning is a wide research field with several distinct approaches. In this paper we focus on \emph{inductive learning}, which is a corner stone in machine learning. 
Even with this specific focus, the amount of relevant research  is vast, and the aim of the survey is not to provide an overview of all published work, but rather to cover the wide range of different usages of the term bias.

This paper is organized as follows. Section~\ref{sec:RelatedWork} briefly summarizes related earlier work.  In Section~\ref{sec:Sources} we survey various sources of bias, as it appears in the different steps in the machine learning process.
Section~\ref{sec:Model} contains a survey of various ways of defining bias in the model that is the outcome of the machine learning process. In Section~\ref{sec:Conclusions} we provide a taxonomy of bias, and discuss the different types of found biases and how they relate to each other. Section \ref{sec:FinalRemarks} concludes the paper.

\section{RELATED WORK} \label{sec:RelatedWork}
A number of reviews, with varying focuses related to bias  have been published recently.
Barocas and Selbst \cite{Barocas14} give as good overview of various kinds of biases in data generation and preparation for machine learning.
Loftus et al.~\cite{Loftus18} review a number of both non causal and causal notions on fairness, which is closely related to bias.
Suresh and Guttag~\cite{Suresh2019AFF} identify a number of sources of bias in the machine learning pipeline.
Olteanu et al.~\cite{Olteanu19} investigate bias and usage of data from a social science perspective. 
Our analysis is complementary to the work cited above, by focusing on bias in conjunction with machine learning, and by examining a wider range of biases. We also provide a novel analysis and discussion on the connections and dependencies between the different types of biases. 

\section{SOURCES OF BIAS} \label{sec:Sources}
Our survey of sources of bias is organized in sections corresponding to the major steps in the machine learning process (see Figure~\ref{fig:model}).
In Section~\ref{sec:BiasInLearning} we describe bias as a major concept in the learning step. 
In Section~\ref{sec:BiasInThe World} we focus on our biased world, which is the source of information for the learning process.
Section~\ref{sec:BiasInDataGen} describes the plethora of biases related terms used in the data generation process.

\subsection{Bias in learning} \label{sec:BiasInLearning}
In inductive learning, the aim is to use a data set $ \{ (x_i,y_i) \}_{i=1}^N$ to find a function $f^*(x)$ such that $f^*(x_i)$ approximates $y_i$ in a good way.
Each $x_i$ is a vector of \emph{features}, while $y$ is denoted the \emph{output}, \emph{target variable} or \emph{label}.
An often-discussed case is when machine learning is used to build decision support systems that outputs recommendations, for example on whether to accept or deny loan applications in a bank. The features may in this case be income, property magnitude, family status, credit history, gender, and criminal record for a loan applicant, while the output $y$ is a recommendation, 0 or 1.

Without further restrictions, infinitely many functions perfectly match any given data set, but most of them are typically useless since they simply memorize the given data set but generalize poorly for other data from the same application. Therefore, the search for a good $f^*$ has to be limited to a certain space $\Omega$ of functions. For example, the function may be assumed to be linear, which is the assumption in linear regression. While this may be sufficient in some cases, more complex function spaces, such as high-order polynomials, or artificial neural networks are often chosen. Each specific function in $\Omega$ is typically defined by a set of numerical weights.

This preference of certain functions over others was denoted \emph{bias} by Tom Mitchell in his paper from 1980 with the title \emph{The Need for Biases in Learning Generalizations} \cite{Mitchell80}, and is a central concept in statistical learning theory. The expression \emph{inductive bias} 
(also known as \emph{learning bias}) is used to distinguish it from other types of biases.

In general, inductive learning can be expressed as the minimization problem 
\begin{equation} \label{eq:min1}
    f^* = \argmin_{f \in \Omega} L(f),
\end{equation}
\noindent
where $L(f)$ is a $cost function$ quantifying how well $f$ matches the data. The most common loss function is defined as
\begin{equation} \label{eq:loss}
    L(f)={\sum_{i=1}^N{(f(x_i)-y_i)^2}}.
\end{equation}

The process of solving the minimization problem in Eq. \ref{eq:min1} is called \emph{training}, and uses a \emph{training set} $ \{ (x_i,y_i) \}_{i=1}^N$ to search through $\Omega$ for the function $f$ that minimizes the loss function in Eq. \ref{eq:loss}. 

All machine learning techniques for inductive learning (for example neural networks, support vector machines, and K-nearest neighbor), need some kind of inductive bias to work, and the choice of $\Omega$ is often a critical design parameter. Having too low inductive bias ($\Omega$ too big) may lead to overfit, causing noise in data to affect the choice of $f^*$. This leads to bad generalization, meaning that $f^*$ does not approximate new data that was not used during learning. On the other hand, having too high inductive bias ($\Omega$ too small) may lead to underfit, meaning that $f^*$ approximates both the used data and new data in an equally poor way. 

The learning step includes other types of bias than the inductive bias described above. Many machine learning algorithms, in particular within deep learning, contain a large number of \emph{hyper-parameters} that are not learned during training but have to be chosen by the user~\cite{bertrand19}. For neural networks, the choice of number of hidden nodes and layers and type of activation functions are strictly part of the definition of $\Omega$, but are often seen as hyper-parameters~\cite{ZhangChen19}. Other hyper-parameters are related to the way the optimization problem in Eq.~\ref{eq:min1} is solved. Besides the choice of algorithm (for example back propagation, Levenberg-Marquardt, or Gauss-Newton), learning rate, batch size, number of epochs, and stopping criteria are all important choices that affect which function $f^*$ is finally chosen.
The shape of the loss function $L(f)$ is another hyper-parameter that not necessarily have to be  as in Eq.~\ref{eq:loss}. Examples of other possible choices are \emph{Mean Absolute Error, Hinge Loss}, and \emph{Cross-entropy Loss}.
The choice of meta-parameters usually have a large impact on the resulting model. We propose to denote this particular type of bias as \emph{hyper-parameter bias}.

The learning step involves more possible sources of bias.
One example is denoted \emph{uncertainty bias}~\cite{Goodman2017EuropeanUR}, and has to do with the probability values that are often computed together with each produced classification in a machine learning algorithm\footnote{For many classification algorithms for example K-nearest neighbor, Artificial Neural Networks, and Na\"{\i}ve Bayes, the output is the probabilities for each class $y_i$, given the input $x$:  $P(y_i \mid x)$. The predicted class is simply the class with the highest such probability.}. The probability represents uncertainty, and typically has to be above a set threshold for a classification to be considered. For example, a decision support system for bank loan applications may reject an application although it is classified as `approve', because the probability is below the threshold. This threshold is usually manually set, and may create a bias against underrepresented demographic groups, since less data normally leads to higher uncertainty.

\subsection{A Biased World} \label{sec:BiasInThe World}

The word `bias' has an established normative meaning in legal language, where it refers to `judgement based on preconceived notions or prejudices, as opposed to  the impartial evaluation of facts'~\cite{campolo2018ai}. The world around us is often described as biased in this sense, and since most machine learning techniques simply mimic large amounts of observations of the world, it should come as no surprise that the resulting systems also express the same bias. 

This bias of the world is sometimes denoted  \emph{historical bias}. Historical bias has to do with data that in itself has unwanted properties that are regarded as biased: `Historical bias arises even if the data is perfectly measured and sampled, if the world \emph{as it is} or \emph{was} leads a model to produce outcomes that are not wanted'~\cite{Suresh2019AFF}. 
The bias of the world obviously has many dimensions, each one describing some unwanted aspect of the world.

Sometimes, the bias in the world is analyzed by looking at correlations between features, and between features and the label. The authors of ~\cite{ZhaoEtAl2017} show that in a certain data set, the label {\em cooking}  co-occurs unproportionally often with {\em woman}, as compared to {\em man}. Since most machine learning techniques depend on correlations, such biases may propagate to learned models or classifiers. 
The authors of \cite{ZhaoEtAl2017} show examples of this, and present techniques to detect and quantify bias related to correlations. They define a subset of output variables $G$
that reflect a demographic attribute such as gender or race
(e.g. $G = \{ man, woman\}$), and 
$o$ as a variable that is potentially  correlated with the elements of $G$ (e.g. $o = {\emph{cooking}}$). To identify unwanted correlations, a bias score for $o$, with respect to a demographic variable $g \in G$, is defined as

$$b(o,g) = \dfrac{c(o,g)}{\sum\limits_{g' \in G} c(o,g')},$$

\noindent where $c(o,g)$ is the number of occurrences of $o$ and $g$ in a corpus. 
If $b(o,g) > 1 / ||G||$, then $o$ is positively correlated with $g$, which indicates that data is biased in this respect. 
To identify this particular notion of bias, we propose using the term \emph{co-occurrence bias}.

A large body of research investigate bias properties of text, at sentence level, paragraph level, article level, or entire corpora such as Wikipedia news. 
In some published work, the word `bias' simply denotes general, usually unwanted, properties of text~\cite{RecasensEtAl2013,hube2018towards}. 
\emph{Framing bias} refers to how a text expresses a particular opinion on a topic. 
The connection between framing bias and gender/race bias is investigated in~\cite{Kiritchenko2018ExaminingGA}, which presents a corpus with sentences expressing negative bias towards certain races and genders. 
Another example of text related bias is \emph{epistemological bias}, which refers to the degree of belief expressed in a proposition. 
For example, the word {\em claimed} expresses an epistemological bias towards doubts, as compared to {\em stated}.

As the authors of \cite{HubFet2018} conclude, text related bias depends not only on individual words, but also on the context in which they appear.
The authors use the broader term \emph{language bias} with reference to to the guidelines for Neutral Point of View\footnote{\href{https://en.wikipedia.org/wiki/Wikipedia: Neutral\_point\_of\_view}{Wikipedia:Neutral point of view}. Accessed July 29, 2020.} (NPOV). Aimed for Wikipedia editors writing on controversial topics, NPOV suggests to `(i) avoid stating opinions as facts, (ii) avoid stating seriously contested assertions as facts, (iii) avoid stating facts as opinions, (iv) prefer nonjudgemental language, and (v) indicate the relative prominence of opposing views'.
Wagner et al.~\cite{Wagner2015} present and apply several measures for assessing gender bias in Wikipedia. \emph{Coverage bias} is computed by comparing the proportions of notable men and women that are covered by Wikipedia (the somewhat surprising result is that this proportion is higher for women). 


\subsection{Bias in Data Generation} \label{sec:BiasInDataGen}
In machine learning,  data generation is responsible for acquiring and processing observations of the real world, and deliver the resulting data for learning.
Several sub-steps can be identified, each one with potential bias that will affect the end result.
As a first step, the data of interest has to be specified. The specification guides the measurement step, which may be automatic sensor based data acquisition, or manual observations of phenomena of interest. For inductive learning, data is then usually manually labelled. In the following, possible sources of bias in each of these sub-steps will be surveyed.

\subsubsection{Specification bias} \label{sec:SpecificationBias}
We propose the term \emph{specification bias} to denote bias in the choices and specifications of what constitutes the input and output in a learning task, i.e.:
\begin{itemize}
 \item The features in the vectors $x_i$ in Eq.~\ref{eq:loss}, for example `income', `property magnitude', `family status', `credit history', and `gender' in a decision support system for bank loan approvals.
 \item The output $y$ in Eq.~\ref{eq:loss}, for example `approve' as target variable. Sometimes the choice of target variable involves creation of new concepts, such as `creditworthiness', which adds extra bias.
 \item In the case of categorical features and output, discrete classes related to both $x$ and $y$, for example `low', `medium', and `high'.
\end{itemize}
These specifications are typically done by the designer of the system, and require good understanding of the problem, and an ability to convert this understanding into appropriate entities \cite{Chapman00}. Unintentionally or intentionally biased choices may negatively affect performance, and also systematically disadvantage \emph{protected classes} in systems building on these choices \cite{Barocas14}. 

Related to the selection of features, the notion of~\emph{proxies} deserves some comments. When designing a decision support system, one  approach to prevent bias with respect to a protected attribute, such as race, is to simply remove  race  from the features used for training. One problem with this approach is that the result may still be biased with respect to race, if other features are strongly correlated with race and therefor act as \emph{proxies} for race in the learning~\cite{DattaFKMS17aa,Suresh2019AFF}. Proxies for race could, for example, be area code, length, and hairstyle.

\subsubsection{Measurement bias} 
In epidemiology, \emph{Measurement bias}, \emph{Observational bias}, and \emph{Information bias} refer to bias arising from measurement errors~\cite{rothman2015modern}, i.e. errors occurring in the process of making observations of the world. In the reviewed material on bias and machine learning, such bias was rarely mentioned, although this process can be biased in very many ways. In epidemiology and medicine, the data gathering process is central, and the Dictionary of Epidemiology~\cite{Porta08} lists 37 different types of biases that may influence data. While most of the listed biases are specific for medicine and epidemiology, we identified the following fundamental types of measurement related bias that are highly relevant also for machine learning. \emph{Bias due to instrument error} is a `Systematic error due to faulty calibration, inaccurate measuring instruments, contaminated reagents, incorrect dilution or mixing of reagents, etc.'. \emph{Observer bias} is defined as `Systematic difference between a true value and the value actually observed due to observer variation'. The related \emph{investigator bias} is defined as `Bias on the part of the investigators of a study toward a particular research result, exposure or outcome, or the consequences of such bias'. Hence, a measurement bias can occur either due to the used equipment, or due to human error or conscious bias.

\subsubsection{Sampling bias} 
Sampling bias occurs when there is an underrepresentation or overrepresentation of observations from a segment of the population~\cite{OnlineStat}. 
Such bias, which is sometimes called \emph{selection bias}~\cite{campolo2018ai}, or \emph{population bias}~\cite{Olteanu19}, may result in a classifier that performs bad in general, or bad for certain demographic groups. One example of underrepresentation is a reported case where a  New Zealand passport robot rejected an Asian man's eyes because `subject eyes are closed'\footnote{\href{https://edition.cnn.com/2016/12/07/asia/new-zealand-passport-robot-asian-trnd/index.html}{CNN World, Dec. 9, 2016}. Accessed July 29, 2020.}. 
A possible reason could have been that the robot was trained with too few pictures of Asian men, and therefor made bad predictions on this demographic group. 

There are many reasons for sampling bias in a dataset. One 
kind is denoted \emph{self-selection bias}~\cite{OnlineStat} and can be exemplified with an online survey about computer use. Such a survey is likely to attract people more interested in technology than is typical for the entire population and therefor creates a bias in data. Another example is a system that predicts crime rates in different parts of a city. Since areas with more crimes typically have more police present, the number of reported arrests would become unfairly high in these areas. If such a system would be used to determine the distribution of  police presence, a viscous circle may even be created~\cite{Cofone17, Rich19}.

An opposite example demonstrates how the big data era with its automatic data gathering can create `dark zones or shadows where some citizens and communities are overlooked'~\cite{Crawford2013ThinkAB}. The author Kate Crawford points to \emph{Street Bump}, a phone app that uses the phone's built in accelerometer to detect and report information about road problems to the city. Due to the uneven distribution of smartphones across different parts of the city, data from Street Bump will have a sampling bias.

It is important to note that sampling bias does not only refer to unbalanced categories of humans, and furthermore not even to unbalanced categories. Unbalances may also concern features that have to appear in a balanced fashion. 
One example is given in \cite{Torralba11}, and is there denoted \emph{dataset bias}. Focusing on image data, the authors argue that `... computer vision datasets are supposed to be a representation of the world', but in reality, many commonly used datasets represent the world in a very biased way. Objects may, for example, always appear in the center of the image. This  bias makes it hard for a classifier to recognize objects that are not centered in the image \cite{Torralba11}. 
The authors compared six common datasets of images used for object detection, and found that performance on another dataset than the one used during training in average was cut to less than half.
A similar effect is reported in~\cite{Bahng2019}. If all images in a dataset containing a snowmobile also contain snow, a machine learning algorithm may find snow cues useful to detect snowmobiles. While this may work fine for images in the dataset used for training, it becomes problematic to analyze images with snowmobiles placed indoors. 

Another kind of sampling bias is \emph{survivorship bias}~\cite{OnlineStat}. It occurs when the sampled data does not represent the population of interest, since some data items `died'. One example is when a bank's stock fund management is assessed by sampling the performance of the bank's current funds. This leads to a biased assessment since poorly-performing funds are often removed or merged into other funds \cite{Malkiel95}.

\subsubsection{Annotator bias}
 \emph{Annotator bias} refers to the manual process of labelling data, e.g. when human annotators assign `approve' or `do not approve' to each $y_i$ to be used to build a classifier for approval of loan applications~\cite{Sap19}. During this process, the annotators may transfer their prejudices to the data, and further to models trained with the data. Sometimes, labelling is not manual, and the annotations are read from the real world, such as manual decisions for real historical loan applications. In such cases bias rather falls into the category historical bias (see Section~\ref{sec:BiasInThe World}). 

 \subsubsection{Inherited bias} \label{sec:InheritedBias}
 It is quite common that tools built with machine learning are used to generate inputs for other machine learning algorithms. If the output of the tool is biased in any way, this bias may be inherited by systems using the output as input to learn other models. One example is if the output of a smile detector based on images is used as input to a machine learning algorithm. If the smile detection is biased with respect to age, this bias will propagate into the machine learning algorithm.
 We suggest the term \emph{inherited bias} to refer to this type of bias.
 The authors of \cite{SunEtAl2019} identify a number of Natural Language Processing  tasks that may cause such inherited bias: \emph{machine translation, caption generation, speech recognition, sentiment analysis, language modelling,} and \emph{word embeddings}. For example, tools for sentiment analysis have been shown to generate different sentiment for utterances depending on the gender of the subject in the utterance \cite{Kiritchenko2018ExaminingGA}. 
 Another example is word embeddings, which are numerical vector representations of words, learned from data \cite{Mikolov2013EfficientEO, pennington2014glove}. Word embeddings are often used as input to other machine learning algorithms, and usually provide a powerful way to generalize since word embeddings for semantically close words are close also in the word vector space. However, it has been shown that several common word embeddings are gender biased. 
 The authors in  
 \cite{BolukbasiEtAl2016} 
 show that word embeddings trained on Google News articles exhibit female/male gender stereotypes to a disturbing extent. For example, in the embedding space, the word `nurse' is closer to `female' than to `male'. The authors in \cite{CaliskanEtAl2017}  propose a method called WEAT (Word Embedding Association Test) to measure such bias. Two sets of so-called {\em target words} (e.g. {\em programmer, engineer, scientist \& nurse, teacher, librarian}) and two set of so-called {\em attribute words} (e.g. {\em man, male \& woman, female}) are considered. The null hypothesis is that there is no difference between the two sets of target words in terms of their relative similarity to the two sets of attribute words. 
 
Methods that reduce this kind of bias in word embeddings have been suggested, and either modify already trained word embeddings \cite{BolukbasiEtAl2016} or remove parts of the data used to train the embeddings~\cite{BrunetEtAl2019}. However,  bias may still remain \cite{gonen19} after applying these methods, and may propagate to  models generated by other machine learning algorithms that rely on word embeddings as input.

\section{ASSESSING MODEL BIAS} \label{sec:Model}
The result from an inductive learning process, i.e. the function $f^*$ in Eq. \ref{eq:min1}), is often referred to as a `model'.  As described in the previous sections, bias may propagate from the biased world, through a biased data generation, to the learning step with its inevitable inductive bias and other biases.  The observed bias of a resulting model is often simply denoted `bias'~\cite{Hardt16,campolo2018ai,Cofone17,Chouldechova2016FairPW}. 
To distinguish this from  other types of bias discussed in this paper, we propose using the term \emph{model bias} to refer to bias as it appears and is analyzed in the final model.
An alternative would be the existing term~\emph{algorithmic bias}~\cite{Danks2017AlgorithmicBI}. However, typical usage of that term usually refers to the societal effects of biased systems~\cite{Panch19}, while our notion of bias is broader. Nevertheless, most suggestions on how to define  model bias statistically consider such societal effects: how classification rates differ for groups of people with different values on a  \emph{protected attribute} such as race, color, religion, gender, disability, or family status \cite{Hardt16}. 
As we will see in the following, classification rates may differ in very many respects, and a large number of bias types have been defined based on the condition that should hold for a model not to have that particular type of bias.
For a binary classifier we can for example require that the \emph{overall misclassification rate} (OMR) is independent of a certain protected attribute $A$ (that takes the values 0 or 1). The corresponding condition for a classifier not being biased in this respect is~\cite{Zafar17}:

\begin{equation} \label{eq:OM}
P(\widehat{Y} \neq y|A=0)=P(\widehat{Y} \neq y|A=1),
\end{equation}

\noindent where $\widehat{Y}$ is the classifier output $f(x)$ (see Eq.~\ref{eq:loss}), and $y$ is the correct classification for input $x$. Both $\widehat{Y}$ and $y$ take the values 0 or 1.
For example, the fact that a person is female ($A=0$) should not increase or decrease the risk of incorrectly being refused, or allowed, to borrow money at the bank.
Several similar conditions can be defined to describe other types of unwanted bias in a classifier model~\cite{Zafar17}:

false positive rate (FPR):
\begin{equation} \label{eq:FPR}
P(\widehat{Y} \neq y|A=0,y=0)=P(\widehat{Y} \neq y|A=1,y=0),
\end{equation}

false negative rate (FNR):
\begin{equation} \label{eq:FNR}
P(\widehat{Y} \neq y|A=0,y=1)=P(\widehat{Y} \neq y|A=1,y=1),
\end{equation}

false omission rate (FOR):
\begin{equation} \label{eq:FRO}
P(\widehat{Y} \neq y|A=0,\widehat{Y}=0)=P(\widehat{Y} \neq y|A=1,\widehat{Y}=0),
\end{equation}

false discovery rate (FDR):
\begin{equation} \label{eq:FDR}
P(\widehat{Y} \neq y|A=0,\widehat{Y}=1)=P(\widehat{Y} \neq y|A=1,\widehat{Y}=1).
\end{equation}

Each one of these equations focuses on that an incorrect $(\widehat{Y} \neq y)$ classification should be independent of $A$ and a specific value of  $\widehat{Y}$ or $y$.
The advantageous classifier output (for example being accepted a loan) is here coded as 1.
A classifier that does not satisfy one of these equations is said to be biased in the corresponding sense\footnote{In practise, the requirement is usually that the left and right hand side of the equation should be approximate equal.}.
For example, a classifier is biased with respect to  FDR if the value of $A$ affects the probability of incorrectly being allowed to borrow money.

A related condition is the \emph{equalized odds}, which appears in the literature with slightly different definitions (see~\cite{Hardt16} and~\cite{Loftus18}). 
In~\cite{Hardt16}, equalized odds is defined by the following two conditions (slightly modified notation):

\begin{equation} \label{eq:EqualisedOdds1}
P(\widehat{Y}=1|A=0, y=0)=P(\widehat{Y}=1|A=1, y=0),
\end{equation}
\noindent and
\begin{equation} \label{eq:EqualisedOdds2}
P(\widehat{Y}=1|A=0, y=1)=P(\widehat{Y}=1|A=1, y=1).
\end{equation}
Note that Eq.~\ref{eq:EqualisedOdds1} is equivalent to FPR in Eq.~\ref{eq:FPR}, and Eq.~\ref{eq:EqualisedOdds2} is equivalent to TPR in Eq.~\ref{eq:FNR}.


Several other indicators of model bias have been proposed. Loftus et al.~\cite{Loftus18} define \emph{Calibration, Demographic Parity/Disparate Impact}, and  \emph{Individual Fairness}.
For a binary classification $\widehat{Y}$, and a binary protected group $A$, demographic parity is defined as follows:
\begin{equation} \label{eq:DemographicParity}
P(\widehat{Y}=1|A=0)=P(\widehat{Y}=1|A=1).
\end{equation}
That is, $\widehat{Y}$ should be independent of $A$, such that the classifier in average gives the same predictions to different groups. If the equality does not hold, this is referred to as \emph{disparate impact}.
An example is a software company that wants to reach a better gender balance among their, mainly male, programmers. By following the principle of demographic parity, when recruiting, the same proportion of female applicants as male applicants are hired. 


Taken all together we  conclude that there is a large number of different types of model biases, each one with its own focus on unwanted behavior of a classifier. Furthermore, many of these biases are related, and it can also be shown that several of them are conflicting in the sense that they cannot be avoided simultaneously \cite{Zafar17,KleinbergMR16,Chouldechova2016FairPW}.
Hence, it is problematic to talk about `fair' or `unbiased' classifiers, at least without clearly defining the meaning of the terms. It can also be argued that a proper notion of fairness must be task-specific \cite{Dwork12}.

\section{TOWARDS A TAXONOMY OF BIAS} \label{sec:Conclusions}
In this section we summarize and discuss the various notions of bias found in the survey, and propose a taxonomy, illustrated in Figure~\ref{fig:model}.

\subsection{Terminology}
\begin{figure*}
\begin{center}
\includegraphics[scale=0.45]{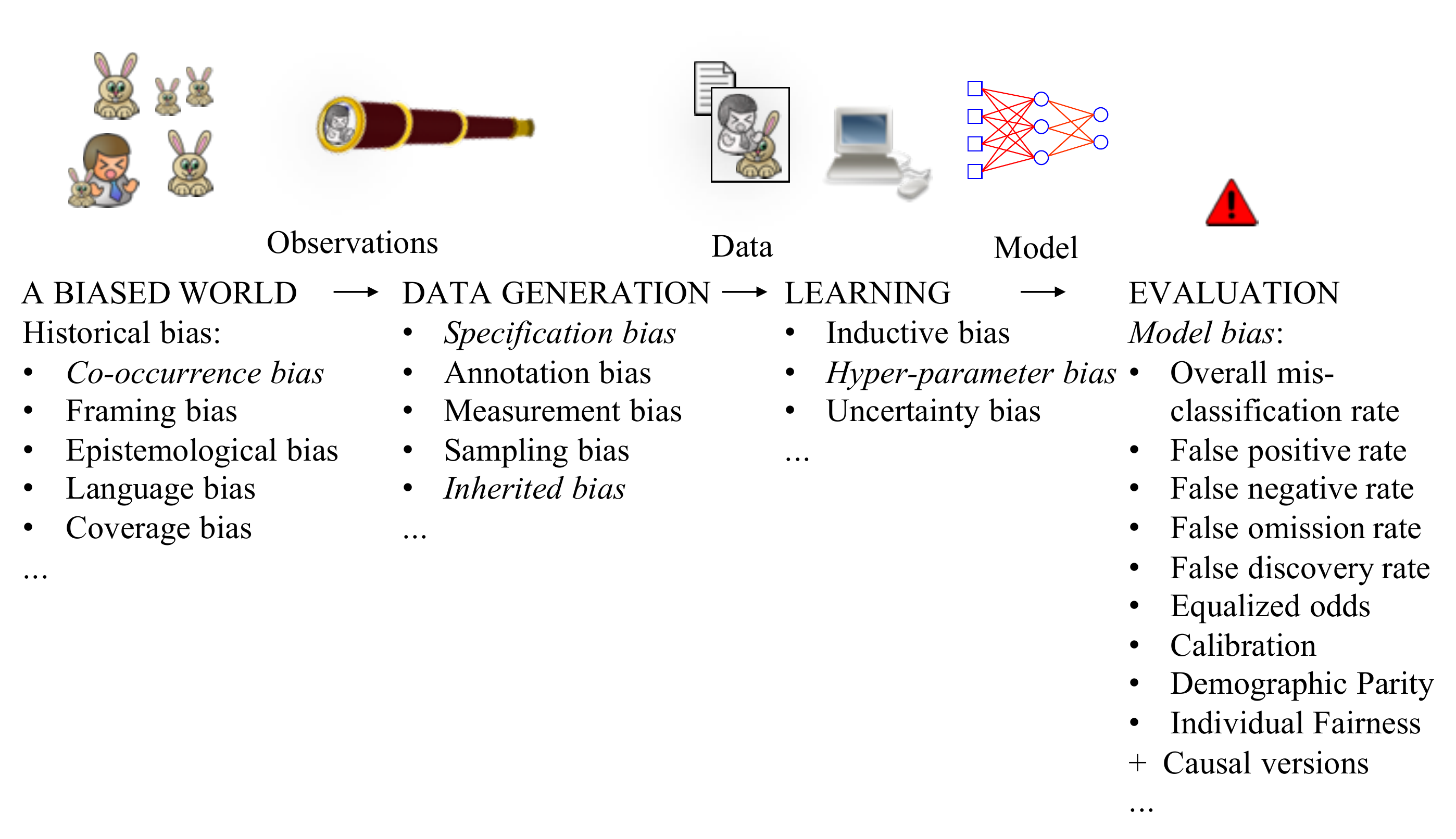}
\end{center}
\caption{Proposed taxonomy of the different types of bias that appear in the machine learning pipeline. 
Starting with existing bias in the world, biased observations are made, and data is  generated in a, possibly biased, process. The data is then used by a biased learning mechanism to produce a model. The bias of the model is finally evaluated. When necessary for clearness or completeness, we propose new terms (here written in italics).
}
\label{fig:model}
\end{figure*}

While it used to be the case that `Bias in machine learning' usually referred to the inductive bias we describe in Section \ref{sec:BiasInLearning}, this is no longer the case. As the survey shows, there is a multitude of usages with different meanings of bias in the context of machine learning. We summarize our proposed taxonomy in Figure~\ref{fig:model}, with different types of biases organized in the three categories \emph{A biased world, Data generation}, and \emph{Learning}. 
In several cases the meaning of terms differed between surveyed papers, and in some cases specific and important types of biases were only referred to as `bias'. In these cases, we propose descriptive names. 

In the Biased world category, the main term is \emph{historical bias}. We identify five named types of historical bias. If we define bias as things that `produce outcomes that are not wanted'~\cite{Suresh2019AFF},  this list could of course be made considerably longer. We suggest the term \emph{co-occurrence bias} for cases when a word occurs disproportionately often together with certain other words in texts (see Section~\ref{sec:BiasInThe World}).

In the Data generation category, we found five types of sources of bias. 
This list should also not be taken as complete, but rather as containing some of the most common and representative examples used in the literature. Several sub-types were also identified (see Section~\ref{sec:BiasInDataGen}). We propose the term \emph{specification  bias} to denote bias in the specifications of what constitutes the input and output in a learning task (see Section~\ref{sec:SpecificationBias}), and we suggest the term \emph{inherited bias} to refer to existing bias in previously computed inputs to a machine learning algorithm (see Section~\ref{sec:InheritedBias}).

In the Learning category, we have the classical inductive bias, but also what we name \emph{hyper-parameter bias}, the bias caused by, often manually set, hyper-parameters in the learning step (see Section~\ref{sec:BiasInLearning}).

We propose using the term~\emph{model bias} to distinguish the bias detected in the outcome of a machine learning system, from the possible reasons for this bias. Specific remarks concerning model bias are presented below.

\subsection{On model bias}
A wast majority of published research refer to social discrimination when talking about bias in machine learning.  A typical, and frequently discussed,  example of such model bias is COMPAS, a computer program used for bail and sentencing decisions. It has been labeled biased against black defendants~\cite{Angwin16}
\footnote{\href{https://www.propublica.org/article/machine-bias-risk-assessments-in-criminal-sentencing}{\emph{Machine Bias - There’s software used across the country to predict future criminals. And it’s biased against blacks}}. ProPublica May 23, 2016. Accessed Feb. 10, 2020.}.

Model bias is caused by bias propagating through the machine learning pipeline. Bias in the data generation step may, for example, influence the learned model, as in the previously described example of sampling bias,  with snow appearing in most images of snowmobiles. This may cause an object classification algorithm to use irrelevant features as shortcuts when learning to recognize snowmobiles (in this case snow cues)~\cite{Bahng2019}.
This in turn leads to a classifier that is biased against snowmobiles placed indoors, and biased for snowmobiles placed outdoors. While this, at first, may not be seen as a case of social discrimination, an owner of a snowmobile shop may feel discriminated against if Google does not even find the shop's products when searching for `snowmobiles'.

In our survey we identified nine aspects of model bias, defined by statistical conditions that should hold for a model not  being biased in a specific way. In addition, several causal versions exist.
Some of the identified conditions are contradictory such that any attempt to decrease one bias will increase another. This is not totally surprising since the conditions are related to common performance measures for classifiers, such as precision and recall, which are known to have the same contradictory relation~\cite[pp. 405]{Hassanien19}.
The contradictory conditions is not a statistical peculiarity, but a very real phenomenon. The COMPAS system mentioned above is indeed biased by certain conditions, but fair by others\footnote{\href{http://www.crj.org/assets/2017/07/9_Machine_bias_rejoinder.pdf}{\emph{False Positives, False Negatives, and False Analyses. Community Resources for Justice}}. Accessed Feb. 15, 2020.}\textsuperscript{,}\footnote{\href{https://www.washingtonpost.com/news/monkey-cage/wp/2016/10/17/can-an-algorithm-be-racist-our-analysis-is-more-cautious-than-propublicas/}{\emph{A computer program used for bail and sentencing decisions was labeled biased against blacks. It’s actually not that clear}}. Washington Post Oct. 16, 2017. Accessed Feb. 15, 2020.}~\cite{Flores16}.

 In some cases, certain types of bias violates intuitive notions of fairness, and may even be prohibited by law. One example is demographic parity (Eq.~\ref{eq:DemographicParity}), which aims at classifiers with the same predictions to different groups. As noted in~\cite{Loftus18}, this  may require positive discrimination, where individuals having different protected attributes are treated very differently. 
 In some cases, this may be a consciously chosen strategy to change societal imbalances, for example gender balance in certain occupations. However, it would probably not be seen as a good idea to apply the same reasoning to correct arrest rates for violent crimes, where men are significantly overrepresented as a group.

Given this complex situation, one should view the different aspects of model bias as dimensions of a multi dimensional concept. They should, together with traditional performance measures, be selected, prioritized and used to guide the design of an optimal, albeit not necessarily statistically `unbiased' machine learning system. As noted in~\cite{Chouldechova2016FairPW},  `... it is important to bear in mind that fairness itself ...  is a social and ethical concept, not a statistical one'.

Most used notions of model bias share a fundamental shortcoming: they do not take the underlying causal mechanism that generated data into account. This is serious not least since the legal system defines discrimination as an identified causal process which is deemed unfair by society~\cite{Hardt16}). Furthermore, the importance of causality in this context is widely recognized among ethicists and social choice theorists~\cite{Loftus18}.
Unfortunately, correlations between observed entities can alone not be used to identify causal processes without further assumptions or additional information.
Several researchers have recently developed causal approaches to bias detection. A causal version of equalized odds, denoted  \emph{Counterfactual Direct Error Rate}, is proposed in~\cite{ZhanBar2018}, together with causal versions of  other types of model biases.  Causal versions of additional types are suggested in~\cite{Loftus18,Hardt16}. Due to space constraints we will not discuss these further, although causal reasoning is seen as critical both for identification and reduction of model bias.

\subsection{The world as it should be vs. the world as it is}
It is important, but not always recognized, that most statistical measures and definitions of model bias, such as Equations~\ref{eq:OM}-\ref{eq:EqualisedOdds2}, use the correct classifications $y$ as baseline when determining whether a model is biased or not.  
If $y$ are observations of humans' biased decisions in the real world (such as historical loan approvals), or humans' biased manual labels created in the data generation process, Eq.~\ref{eq:OM} could be perfectly satisfied, which may be interpreted as the model being free of bias (with respect to overall misclassification rate). However, a more correct interpretation would be that the model is no more, or less, biased than the real world. Assessing the `true' degree of biasedness of a model, requires a notion of an ideal `world as it should be', as opposed to the observed `world as it is'. 
Demographic parity (Eq.~\ref{eq:DemographicParity}) has such a notion built in, namely that the classifier output should be independent of the protected attribute.

There are at least two fundamentally different approaches to address the problem with a biased model. We may debias the computed model, based on an understanding of what `the world as it should be' looks like. For example, word embeddings may be  transformed such that the distance between words describing occupations are equidistant between gender pairs such as `he' and `she'  \cite{BolukbasiEtAl2016}. 
Another approach to address biased models is to debias the data used to train the model, for example by removing biased parts, such as suggested for word embeddings~\cite{BrunetEtAl2019}, by oversampling~\cite{geirhos2018imagenettrained}, or by resampling~\cite{Li2019REPAIRRR}. Debiasing input data can be seen as a technical introduction of (good) bias in the data generation process, but it can also be seen as an attempt to model an ideal `world as it should be' rather than the biased `world as it is'.

The distinction between these two worlds is related to when a specific model is useful or not. If the model is going to be used to predict `the world as it is', model bias may not be problem. Such a model may, for example, be used to predict whether a given loan application will be accepted or not by the bank. Good predictions should model also biased decisions made by the bank. 
On the other hand, if the model is going to be used in a decision support system, we may want it to mimic `the world as it should be', and bias is then highly relevant to detect and avoid in the design of the system.
One example of how `the world as it should be' is chosen as norm, is Google's image search algorithm. Since only 5\% of Fortune 500 CEOs were women (2018), a search for `CEO' resulted in images of mostly men. Since then, Google has reportedly changed the algorithm to display a higher proportion of women~\cite{Suresh2019AFF}.

\subsection{What's wrong with discrimination?}
The necessity of inductive bias in machine learning was mentioned in Section~\ref{sec:BiasInLearning}. The same holds at the level of human learning, as discussed in the area of philosophical hermeneutics~\cite{Hildebrandt19}.
In~\cite{Gadamer75} the author argues that we always need some form of prejudice (or bias) to understand and learn about the world. 
Returning to the example in Section~\ref{sec:BiasInLearning}, a decision support system for approval of bank loans is sometimes described as biased and discriminating if it treats certain groups of people differently. It is important to realize that this difference in treatment, in a general sense, is inevitable and rather the main purpose of such a decision support system: to approves some people’s applications, and reject others\footnote{In machine learning this general ability to distinguish between varying input data is even called `discrimination', but without any negative connotations (see for example~\cite{Kovacs2012}).}. For example, it may be the bank’s policy to not approve applications by people with very low income. While this technically is the same as rejecting people based on ethnicity, the former may be accepted or even required, while the latter is often referred to as `unwanted' \cite{Hardt16}, `racial' \cite{Sap19}, or `discriminatory' \cite{Chouldechova2016FairPW,Pedreshi08} (the terms \emph{classifier fairness} \cite{Dwork12,Chouldechova2016FairPW, Zafar17} and \emph{demographic parity} \cite{Hardt16} are sometimes used in this context).
The difference between features such as `income' and `ethnicity' has to do with the, already cited, normative meaning of the word bias expressed as `an identified causal process which is deemed unfair by society'~\cite{campolo2018ai}. This is further reflected in the notions of \emph{protected groups} and \emph{protected attributes}~\cite{Hardt16}, which simply define away features such as `income', while including features that are viewed as important for equal and fair treatment in our society.

\subsection{Fighting bad bias with good}
With the possible exception of inductive bias, the various types of biases described in this paper are usually used with negative connotations - to describe unwanted behavior of a machine learning system. However, several of the types of biases described are not necessarily bad. 
For example, some kind of specification bias is necessary to setup a machine learning task. The alternative would be to observe everything observable in the real world, which would make learning extremely hard, if not impossible. The choice of features to include in the learning constitute a (biased) decision, that may be either good or bad from the point of view of the bias of the final model.

Likewise, annotator bias is usually regarded as a bad thing, where human annotators inject their prejudices into the data, for example by rejecting loan applications in a way that  discriminates members of a certain demographic group. However, there is of course also a possibility for the human annotators, to consciously or unconsciously, inject `kindness' by approving loan applications by the same members `too often'. Depending on the context, this could be described as a good annotator bias.


%

Increasing the inductive bias in the learning step can even be shown to be a general way to reduce an unwanted model bias. Imposing requirements on $f$, such as Eq. \ref{eq:OM}, can be expressed as constrained minimization \cite{Zafar17} in the inductive learning. Eq.~\ref{eq:min1} may be rewritten as 
\begin{equation} \label{eq:min2}
    f^* =  
        \argmin_{
        \substack{
         \text{s.\,t.}\, f \in \Omega ,\\
         {P(f(x) \neq y|A=0)}= \\
         {P(f(x) \neq y|A=1)}
       }
     }
    {\sum_{i=1}^N{(f(x_i)-y_i)^2}}.
\end{equation}

While the minimization problems \ref{eq:min1} and \ref{eq:min2} seem to be identical, the latter is unfortunately much harder to solve. The constraints are non convex, as opposed to the normal concave case which can be solved by several efficient algorithms. The authors in \cite{Zafar17} approximate the additional constraints such that they can be solved efficiently by convex-concave programming \cite{Shen16}.

However, the imposed requirements on $f$ can also be seen as unconstrained minimization over a restricted function space $\Omega'$
\begin{equation} \label{eq:min3}
    f^* = \argmin_{f \in \Omega', } {\sum_{i=1}^N{(f(x_i)-y_i)^2}},
\end{equation}
where $\Omega'$ is the original $\Omega$, with all functions not satisfying the imposed requirements removed. Hence, in order to decrease unwanted (bad) model bias, we increase the inductive (good) bias by restricting the function space $\Omega$ appropriately. 

\section{FINAL REMARKS} \label{sec:FinalRemarks}
Our survey and resulting taxonomy show that `bias' used in conjunction with machine learning can mean very many different things, even if the most common usage of the word refers to social discrimination in the behavior of a learned model.
Even this specific meaning of the word deserves careful usage, since it comes in a variety of types that sometimes even contradict each other.
Regarding bias in the steps leading to a model in the machine learning pipeline, it may or may not influence the model bias, in a sometimes bad, and sometimes good way.

A final remark is that humans are deeply involved in all parts of the machine learning process illustrated in Figure~\ref{fig:model}:
the biased world,  the data generation process, the learning, and the evaluation of bias in the final model. \emph{Cognitive biases} are systematic, usually undesirable, patterns in human judgment and  are  studied in psychology and behavioral economics. They come in a large variety of shades, and the Wikipedia page\footnote{\href {https://en.wikipedia.org/wiki/List_of_cognitive_biases}{Wikipedia \emph{List of cognitive biases}}. Accessed July 29, 2020.} 
lists more than 190 different types. 
Only a small number of them are directly applicable to machine learning, but the size of the list suggests caution when claiming that a machine learning system is `non-biased'.

\bibliographystyle{ecai}
\bibliography{ecai}

\end{document}